\documentclass{article}

\usepackage{arxiv}

\usepackage[utf8]{inputenc} 
\usepackage[T1]{fontenc}    
\usepackage{hyperref}       
\usepackage{url}            
\usepackage{booktabs}       
\usepackage{amsfonts}       
\usepackage{nicefrac}       
\usepackage{microtype}      
\usepackage{lipsum}
\usepackage{graphicx}
\usepackage{subfiles}
\usepackage{calrsfs}
\usepackage{verbatim}
\usepackage{subfig}
\usepackage{geometry}
\usepackage{wrapfig}
\usepackage{multicol}
\usepackage{multirow}

\title{ABOships - an inshore and offshore maritime vessel detection dataset with precise annotations}

\author{
Bogdan Iancu \\
  Faculty of Science and Engineering\\
  \AA bo Akademi University\\
  20500 Turku, Finland \\
  \texttt{bogdan.iancu@abo.fi} \\
   \And
Valentin Soloviev \\
  Faculty of Science and Engineering\\
  \AA bo Akademi University\\
  20500 Turku, Finland \\
  \texttt{valentin.soloviev@abo.fi} \\
  \AND
Luca Zelioli \\
   Faculty of Science and Engineering\\
  \AA bo Akademi University\\
  20500 Turku, Finland \\
  \texttt{luca.zelioli@abo.fi} \\
    \And
Johan Lilius \\
   Faculty of Science and Engineering\\
  \AA bo Akademi University\\
  20500 Turku, Finland \\
  \texttt{johan.lilius@abo.fi} \\
  
}

\begin{document}
\maketitle
\begin{abstract}
Availability of domain-specific datasets is an essential problem in object detection. Maritime vessel detection of inshore and offshore datasets is no exception, there is a limited number of studies addressing this need. For that reason, we collected a dataset of images of maritime vessels taking into account different factors: background variation, atmospheric conditions, illumination, visible proportion, occlusion and scale variation. Vessel instances (including 9 types of vessels), seamarks and miscellaneous floaters were precisely annotated: we employed a first round of labelling and subsequently, we used the CSRT~\cite{lukezic2017discriminative} tracker to trace inconsistencies and relabel inadequate label instances. Moreover, we evaluated the the out-of-the-box performance of four prevalent object detection algorithms (Faster R-CNN~\cite{ren2015faster}, R-FCN~\cite{R-FCN}, SSD~\cite{SSD} and EfficientDet~\cite{tan2020efficientdet}). The algorithms were previously trained on the Microsoft COCO dataset. We compare their accuracy based on feature extractor and object size. Our experiments show that Faster R-CNN with Inception-Resnet v2 outperforms the other algorithms, except in the large object category where EfficientDet surpasses the latter.  
\end{abstract}


\section{Introduction}
Maritime vessel detection from waterborne images is a crucial aspect in various fields involving maritime traffic supervision and management, marine surveillance and navigation safety. Prevailing ship detection techniques exploit either remote sensing images or radar images, which can hinder the performance of real-time applications \cite{shao2019saliency}. Satellites can provide \textit{near-real} time information, but satellite image acquisition   however can be unpredictable, since it is challenging to determine which satellite sensors can provide the relevant imagery in a narrow collection window \cite{liu2014remote}. Hence, seaborne visual imagery can tremendously help in essential tasks both in civilian and military applications, since it can be collected in real-time from surveillance videos, for instance. 

Ship detection in a traditional setting depends extensively on human monitoring, which is highly expensive and unproductive. Moreover, the complexity of the maritime environment makes it difficult for humans to focus on video footage for prolonged periods of time \cite{shao2018seaships}.
Machine vision, however, can take the strain from human resources and provide solutions for ship detection. Traditional methods based on feature extraction and image classification, involving background subtraction and foreground detection, as well as directional gradient histograms, are highly affected by datasets exhibiting challenging environmental factors (glare, fog, clouds, high waves, rain etc.), background noise or lighting conditions. 

Convolutional neural networks (CNNs) contributed massively to the image classification and object detection tasks in the past years~\cite{deng2009imagenet,druzhkov2016survey,zhang2019recent,wu2020recent,liu2020deep}. They incorporate feature extractors and classifiers in multilayer architectures, whose number of layers regulate the selectiveness and feature invariance.  CNNs exploit convolutional and pooling layers extracting local features, and gradually advancing object representation from simple features to complex structures, across multiple layers. CNN-based detectors can subtract compelling distinguishable features automatically unlike more traditional methods which use predefined features, manually selected. However, integrating ship features into detection proves to be challenging even in this context, given the complexity of environmental factors, object occlusion, ship size variation, occupied pixel area etc. This often leads to unsatisfactory performance of detectors on ship datasets. 

To address ship detection in a range of operating scenarios, including various atmospheric conditions, background variations and illumination, we introduce a new dataset consisting of 9880 images, and annotations comprising $41967$ carefully annotated objects. 

The paper is organized as follows. Section~\ref{sec:related} describes related work, including notable results in  vessel detection and maritime datasets comprising waterborne images.  Section~\ref{sec:dataacq} describes the data acquisition and dataset diversity. In Section~\ref{sec:design},  we discuss dataset design and our relabelling algorithm along with basic dataset statistics based on the final annotation data. In Section~\ref{sec:basline}, we discuss evaluation criteria and present experimental results; we investigate 4 CNN-based detectors and discuss the feature extractors and object size effect on the performance of the detectors. Conclusions are presented in Section~\ref{sec:conclusion}.

\section{Related Work}\label{sec:related}
\subsection{Object Detection}
Object detection is one of the fundamental visual recognition problems where the requirement is to predict whether there are any objects from given categories in an image and provide their location (bounding box or pixel-level localization in case of instance segmentation), if any are found. Generally, this is achieved by extracting features in an image and matching them against features from trained images. Traditional approaches use sliding window to generate proposals, then visual descriptors to generate an embedding, which are subsequnetly classified (such as SVM, bagging, cascade learning and adaboost). Traditional algorithms with best performance focus on carefully designing the descriptors for extracting the features (SIFT, Haar, SURF), however, starting the year 2008, more and more limitations of this approach became evident\cite{wu2020recent}. We list below the most notable ones:
\begin{itemize}
    \item Hand annotated visual descriptors provided large number of proposals  which caused high amount of false positives. 
    \item Visual descriptors (as mentioned above) extract low-level features, but are unsuitable for high-level features.
    \item Each step of detection pipeline is optimized separately, so global optimization is unattainable.
\end{itemize}

In early 2010s, deep learning approaches came to prominence and started replacing the traditional ones. The object detection networks can be roughly categorized into 2 types: one-stage detectors and two-stage detectors. The structure of the latter resembles traditional object detectors in that it generates proposal-regions and then classifies the proposals, while the former considers positions of an image as potential objects and tries to classify them immediately. The traditional approach of sliding windows for proposal generation is still used in CNNs, but there have been other notable advances that allow for more efficient proposal generation such as anchor-based and key-point approaches (CenterNet being one of the more notable examples of the kind)~\cite{wu2020recent}. 

However, the key difference between traditional object detection and CNNs is in the way that visual descriptors are generated. In deep learning, instead of creating visual descriptors by hand, the same role is performed by convolutional layers. Instead of defining feature extractors by hand, basic CNNs train multiple convolutional layers to extract both high- and low-level features, which are then classified with the help of fully-connected layers. The resulting network essentially solves all the main limitations of a traditional approach, but the trade-off is that the resulting network requires a significantly larger number of training images for optimizing all parameters~\cite{wu2020recent,liu2020deep}.

While the requirement of a large number of training samples can prove to be a large obstacle, one of the benefits of CNNs is that models involving them can be generalized into other fields with similar characteristics with the help of transfer learning.  By training a model on a specific dataset, the backbone of the model can be later used to extract features in other tasks with similar features. For this reason, the aim of recent CNN-models was to be as generic as possible, since with the help of transfer learning, they can be specialized for whatever field is needed. The challenge, however, is when the generic models are not suitable feature extractors for a new field and there is not enough data to train them~\cite{zhang2019recent}. For those specific cases, the only solution is creation of new datasets.

\subsection{General Object Detection Datasets}
The two main reasons for the remarkable progress computer vision made in the past decades are the availability of large-scale datasets and powerful GPUs that made it possible for deep learning to take off considerably~\cite{pathak2018application}. Deep learning has made notable contributions to the field of computer vision, image classification and object detection being at the forefront of the research areas that benefited from it. International competitions such as ILSVRC, PASCAL VOC, and Microsoft COCO motivated the community to make tremendous contributions, each of them offering large-scale datasets that have been exploited ever since. These general object detection datasets have been extensively used for object detection with deep neural networks. They are essential for testing and training computer vision algorithms. We will discuss below some of the most prominent general purpose object detection datasets. 

Microsoft COCO~\cite{lin2014microsoft} provides a selection of 328000 images with a number of 2.5 million of labelled object instances, over 91 object classes. The dataset labeling used per-instance segmentation to ensure precise object localization. Two crucial aspects of the dataset are that it exhibits abundant contextual information and images contain multiple objects per image. 

The ImageNet Large Scale Visual Recognition Challenge (ILSVRC) ran annually for a number of years and was established as one of the typical benchmarks for object classification and detection. The Imagenet dataset~\cite{deng2009imagenet}, the foundation of the challenge, is an image collection based on the WordNet hierarchy~\cite{miller1995wordnet}, which provides on average 1000 images of manually verified images for every \textit{synset} (synonym set) in the hierarchy. These images are subjected to quality-control and are human-annotated. The dataset consists of over 14 million images, of which over 14 million were annotated to denote what objects are present in the image and, for at least a million of them, bounding boxes are provided too. 

Pattern Analysis, Statistical Modelling and Computational Learning (PASCAL) Visual Object Classes (VOC) is a prominent project in the computer vision community, which provided publicly available image datasets including ground truth annotations and standardized evaluation metrics. These datasets were exploited as part of a number of challenges on various tasks such as: classification, detection, segmentation, etc. The greater number of scientific publications regarding object detection use the PASCAL VOC challenges to benchmark their proposed algorithms. The reason for that is that these challenges introduced a number of evaluation methods: bootstrapping, to decide significant diferences among algorithms, a normalised average precision across classes, etc. The dataset released by last PASCAL VOC challenge includes 11530 images with 27450 annotated objects over 20 classes. Table~\ref{GeneralDatsets} shows a variety of object detection datasets, with their total number of images and clasess. We can notice that ImageNet is by far the largest of the ones mentioned in the table, encompassing the greater number of total images and classes.

\begin{table}[htb]
    \centering
    \begin{tabular}{ |p{4cm}|p{3cm}|p{3cm}|p{3cm}| }
\hline
\multicolumn{4}{|c|}{General Object Detection Dataset} \\
\hline
Dataset & Total images & Total classes & Annotations \\
\hline
ImageNet  & 14000000 & 1000 & 1034908  \\
COCO      & 328000   & 91   & 2500000   \\
OpenImage (6) & 9000000  & 600  & 16000000  \\
PASCAL VOC (2012)   & 11530    & 20   & 27450    \\
\hline
\end{tabular}\bigskip
    \caption{Different object detection datasets comprising various object classes, with their corresponding annotations.}
    \label{GeneralDatsets}
\end{table}

Of the general-purpose object detection datasets, in Table~\ref{GeneralDatsets}, the total number of maritime vessel included are limited, only Microsoft COCO comprising a considerable amount of vessels, $3146$. All vessel counts can be found in Table~\ref{GENDSET:SHIPS}. 

\begin{table}[htb]
    \centering
    \begin{tabular}{|p{2cm}|p{2cm}|}
    \hline
\multicolumn{2}{|c|}{Maritime vessel instances} \\
\hline
Dataset & Vessel count \\
\hline
ImageNet   & 1071 \\
COCO       & 3146 \\
OpenImage  & 1000 \\
PASCAL VOC &  353 \\
\hline
\end{tabular}\bigskip
    \caption{Maritime vessel instances in general object detection datasets.}
    \label{GENDSET:SHIPS}
\end{table}

\subsection{Maritime Vessel Detection Datasets}
Maritime vessel detection from satellite imagery was employed in many studies, over the past 40 years, a review from 2018, \cite{kanjir2018vessel}, gathering a number of 119 papers regarding ship detection and classification only from optical satellites. At the same time the studies regarding maritime vessel detection from waterborne images are still quite scarce to this day. Some studies proposed algorithms utilizing the idea of background subtraction and detection of the foreground in maritime images. This class of techniques is predominantly used in surveillance applications due to their ability to perform well with unexpected changes in illumination, frequency or background noise, \cite{arshad2010multiple}. Other studies proposed solutions for ship detection based on the Histogram of Oriented Gradients (HOG) and sliding windows \cite{wijnhoven2010online}. 

However, since the bloom of deep learning in the past 15 years, CNNs were employed in ship detection from waterborne images. Even so, datasets of seaborne images are scarce, the most notable ones we briefly discuss below. 

The Singapore Maritime Dataset, introduced in \cite{lee2018image} consists of 80 videos recorded during daytime and nighttime, and provides ground truth labels for every frame of every video, comprising bounding-boxes and object classes for the corresponding bounding-box. The annotations for the Singapore Maritime Dataset include 10 classes of ships. This dataset is used for ship detection employing the YOLO v.2 algorithm \cite{redmon2017yolo9000}.

Another recent ship dataset, SeaShips \cite{shao2018seaships}, consists of over 31455 inshore and offshore images of ships, comprising 6 ships types. In \cite{shao2018seaships}, they employ three object detectors (Faster R-CNN \cite{ren2015faster}, SSD \cite{SSD}  and YOLO \cite{redmon2017yolo9000}) to detect ships. 

One of the most recent datasets published is MCShips \cite{zheng2020mcships}, comprising a number of 14709 images of ships, whose annotations cover 6 warship classes and 7 civilian ship classes. In \cite{zheng2020mcships}, they also use the object detection algorithms above (Faster R-CNN \cite{ren2015faster}, SSD \cite{SSD}  and YOLO \cite{redmon2017yolo9000}) to evaluate the dataset over the 13 ship classes. 

We compared our ABOships dataset against other existing ship datasets. Table~\ref{tab:shipdet_datasets} illustrates the main differences. Our datasets has the smallest number of images ($9880$) amongst the four datasets, however it contains a great number of annotations ($41967$) given the images total, which shows it represents well real scenarios of maritime imagery given there are on average more than 4 annotated objects per image.  

\begin{table}[htb]
    \centering
    \begin{tabular}{ |p{1.6cm}|p{1.6cm}|p{1.6cm}|p{1.6cm}| }
    \hline  
    \multicolumn{4}{|c|}{Datasets for ship detection} \\    
    \hline
    Name & Total images & Annotations & Ship types included \\
    \hline  
    SeaShips & 31455 & 40077 & 6 \\
    Singapore  & 17450 & 192980 & 6\\
    MCShips & 14709 & 26529 & 13 \\
    ABOShips & 9880 & 41967 & 9\\
    \hline
    \end{tabular}\bigskip
    \caption{Comparison of ABOships with other maritime datasets.}
    \label{tab:shipdet_datasets}
\end{table}

\section{Data Acquisition}\label{sec:dataacq}
\subsection{Camera System}
The dataset was acquired from a set of 135 videos, collected from a sightseeing watercraft by a camera  with a field of view of $65^{\circ}$ and stored in FullHD ($1920$x$720$) resolution at $15$ FPS in MPEG format. The route of the watercraft extended from the city of Turku to Ruissalo in South-West Finland, the videos comprising the urban area along the Aura river, the port and into the Finnish Archipelago, for a duration of 13 days (26.6.2018-8.7.2018). The watercraft ran each day in the timeframe between 10.15 and 16.45. 
The videos were captured into 30-minute long periods from the route that the watercraft took. While the route remained largely the same, the data contains a variety of typical maritime scenarios in a range of weather conditions.

In addition to camera video data, the platform had a lidar attached to it (Sick LDMRS, FoV $110$ degrees, $2$x$4$ planes, up to $300m$ detection, at $5Hz$). The data from the lidar was captured alongside the video at a rate of $5$ entries of up to $800$ points per second. Our lidar had a detection range of up to $300$ meters, which is very useful for detecting other objects in the harbor environment. Due to having only $2$ times $4$ lasers in height diretion, the data is not reliable enough for discerning what object is detected. It does allow though for detecting distances to the objects seen in the video. For the purpose of creating the dataset presented in this paper, we used the lidar data to filter out video segments that were captured in the harbor area (usually the ones that had too many points for a prolonged period of time).

To evaluate the models, we acquired 9880 image photos from the videos. First, we annotated all images with 11 categories: seamarks, 9 types of maritime vessels, and miscellaneous floaters. 
In a second round, we relabelled all the inconsistencies we found, using an algorithm based on the CSRT tracker~\cite{lukezic2017discriminative}.

\subsection{Dataset Diversity}
Maritime environments are inherently intricate, hence a range of factors have to be accounted for when desinging a dataset. Dataset design must ensure that the dataset characterizes well vessels in the environment. Of course,  data augmentation methods can be considered for reproducing certain environmental conditions, however authentic conditions may be difficult to anticipate.  

\begin{figure}[!htb]
  \centering
    \centering
    \includegraphics[width=1\linewidth]{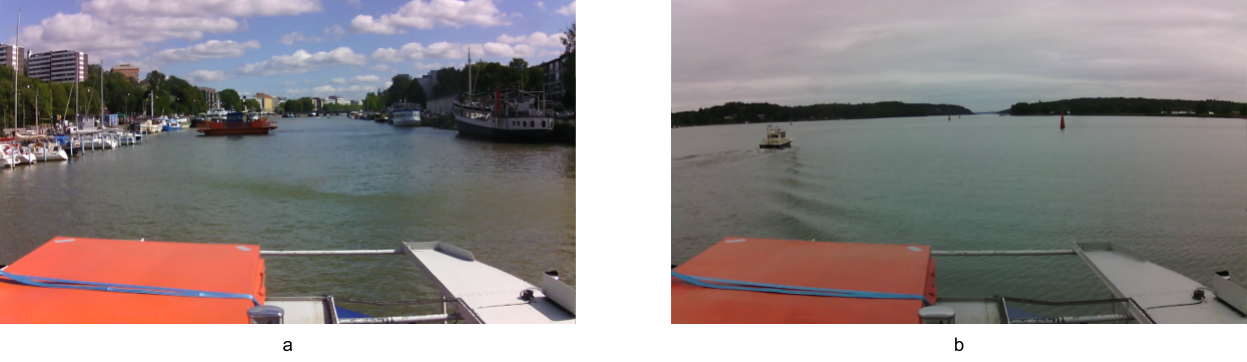}
    \caption{Example image of background variations in the ABOships dataset: (a) View of maritime vessels on Aura river including the urban landscape; (b) View of maritime vessel in the Finnish Archipelago.}
    \label{fig:label_backgroundvariation}
\end{figure}

\textit{Background variation.}
Particular object detection tasks are more prone to to be affected by changes in the background of the picture. For instance, facial recognition is less susceptible to background variations, because given the similar shape of most faces, it is easier to fit them into bounding boxes in a congruous manner. However, the shapes of maritime vessels are highly heterogeneous, making them more difficult to separate from the background due to a potentially vast background information in the bounding box. The accuracy of ship detection would be significantly affected if background information were classified as ship features. 
Figure~\ref{fig:label_backgroundvariation} illustrates the background variation of images in our datasets, including urban landscapes and an opensea environment. 

\textit{Atmospheric conditions.} 
Atmospheric conditions were specific to Finnish summers, with very sunny periods, alternating with rainy intervals and cloudy skies. The dataset includes a variety of images of different atmospheric conditions throughout a day.

\textit{Illumination.}
Lighting variations can significantly impact image capture. Illumination throughout the day, in various geographical areas and with specific daylight hours in a given region can dramatically influence image detection. 

\textit{Visible proportion.}
A great number of the images in our dataset consists of moving ships, with objects being only partially captured in the camera field of view. However, they still represent objects that were annotated since one has to detect them as well. The annotation should comprise different visible proportions of the maritime vessels.    

\begin{figure}[!htb]
  \centering
    \centering
    \includegraphics[width=1\linewidth]{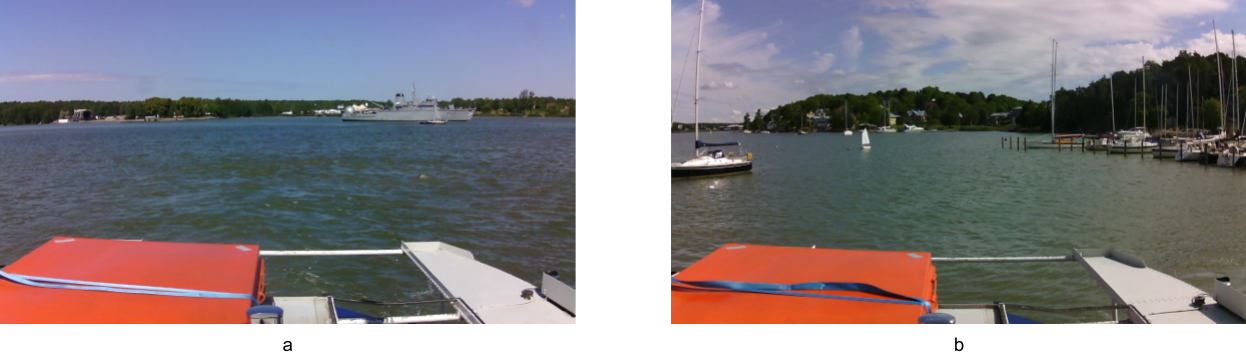}
    \caption{Example image of occlusion: (a) Boat in front of a militaryship; (b) Several sailboats occluding each other while docked, on the right half of the image.}
    \label{fig:label_occlusion}
\end{figure}

\textit{Occlusion.}
Due to the fact that our dataset has been captured in an open sea environment, in the harbor area and also comprises urban landscapes, there are many occasions when maritime vessels occlude each other or occlude other objects in the environment in the harbor area or in the urban landscape. In a subset of pictures especially in the harbor area, there is significant occlusion due to a high number of maritime vessels in the proximity of each other. Two examples of occlusion are shown in Figure~\ref{fig:label_occlusion}.

\textit{Scale variation.} 
Detection of small objects can prove to be quite difficult, especially in a complex environment like the sea, ships that occupy a small pixel area in the picture can be confused with other objects in the background. Maintaining a high level of detection for ships requires including several scales for ships sizes in the dataset. For more information regarding the annotation and the size of the bounding boxes, please refer to Section \ref{subsec:annotation}.

\begin{figure}[!htb]
  \centering
    \centering
    \includegraphics[width=1\linewidth]{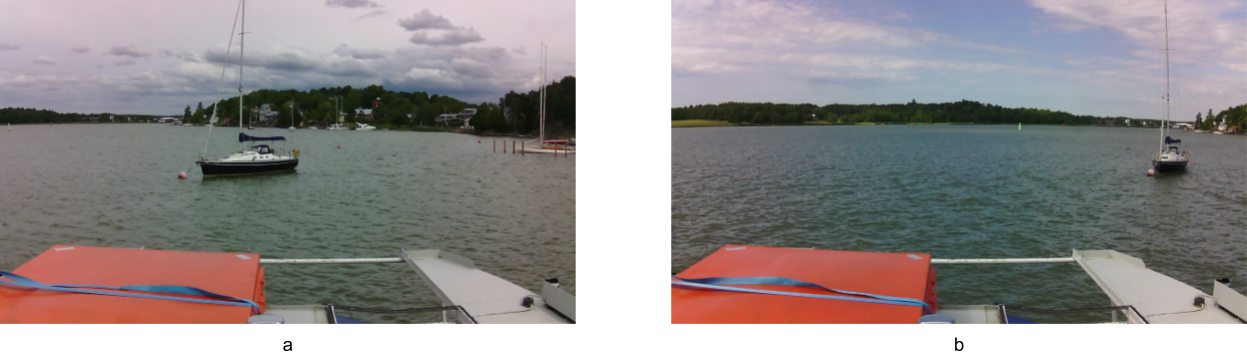}
    \caption{Example image of a sailboat, view from two perspectives: (a) Lateral; (b) Frontal.}
    \label{fig:label_perspective}
\end{figure}

Figure~\ref{fig:label_perspective} illustrates a saiboat from two different perspectives: a lateral and a frontal view, which shows a variation in both occupied pixel area, but also the visible proportion. 

\section{Dataset Design}\label{sec:design}
\subsection{Dataset Design}\label{subsec:design}
The raw data acquired from the camera on the sightseeing watercraft is captured at a resolution of $720p$, as $15$ FPS MPEG videos. The videos include some footage exhibiting content that is irrelevant for the scope of vessel detection (especially footage captured when the vessel was docked, either at the start of its route on the Aura river or at the Port of Turku) or sensitive content, such as faces of people. To address the latter issue, we performed face detection on all videos and blurred all detected faces. Addressing the former issue on the other hand, required additional data from the lidar.

In a maritime environment, lidar data is relatively sparse, authors of this study observed a high number of points detected for a prolonged duration correlates with the vessel being docked in the harbor. By setting a point threshold to detect these docked-harbor cases, we were able to filter them out in their majority and extract only the images from maritime environment. The images were extracted at an interval of 15 seconds (one image every 225 frames) and still contained some images captured during docking, but most of them were facing outwards from harbor, so the images captured that way still contain useful maritime data. As a result we acquired 9880 images of the maritime environment.

The acquired images were subsequently separated into work packages in such a way that chronologically adjacent pictures were separated into different work packages, that were then manually labelled by different people. After the initial labelling was completed, we used the CSRT tracker~\cite{lukezic2017discriminative} to combine labels of the same object into traces - a collection of chronologically adjacent images containing a bounding-box for that object. Due to inaccuracies in the tracking process and discrepancies in labelling, the produced traces were not always accurate. After viewing the labels in these traces, we identified  the main causes for discrepancies in labelling, which were predominantly caused by different interpretations of label annotations. We refined those annotations to eliminate the discrepancies and separated the data into a second collection of work packages that were provided to annotators, who then relabelled the data, according to refined annotations. 
After the relabelling was completed, images were again separated into batches of 20 chronologically subsequent images.

\subsection{Annotation}\label{subsec:annotation}

To perform the annotation task, we first investigated the captured videos and identified the vessel types that appeared most often. Due to the fact that the videos were captured at locations with a significant number of passenger ships, there is a certain level of bias for labellers towards those types of ships. This is different from the Seaships database, for instance, which comprises a higher variety of cargo ships. For the purposes of future use in machine vision, rather than using maritime terminology as such, (depicting ship scale and purpose), we selected labels that had some clearly distinct visual characteristics.

The label categories are discussed below, with more specific details for every category:
\begin{itemize}
    \item seamark (red/blue/black/yellow cone-shaped metal/plastic floater or pipe emerging from the sea);
    \item motorboat - primarily a speedboat, visual distinction - sleek, aerodynamic features;
    \item sailboat - sails-propelled boat or a boat which exhibits sails, visual distinction - sails;
    \item passengership, medium size ship, used to transport people on short distances, ex. restaurant boat,  visual distinction - usually it has multiple noticeable lateral windows;
    \item cargoship, large-scale ships used for cargo transportation, visual distinction - long ship with cargo containers or designed with container carrying capacity;
    \item military ship, an official ship that is either military or coast guard and includes a special hull with antennas. The coast guard usually reads "Coast guard" and the military ones are dark gray/metallic/black/brown in colour;
    \item ferry, medium-sized ship, used to transport people and cars, aka waterbus/watertaxi, another appropriate term would be cableferry, visual distinction - it includes entrances on two opposite sides and a cabin in the middle;
    \item miscboat - miscellaneous maritime vessel, visual distinction - generic boat that does not include any visual distinction mentioned above in the other ship categories above;
    \item miscellaneous - identified floaters(birds, other objects floating in the water) unidentified/unidentifiable floaters.
\end{itemize}

\begin{figure}[!htb]
  \centering
    \centering
    \includegraphics[width=0.7\linewidth]{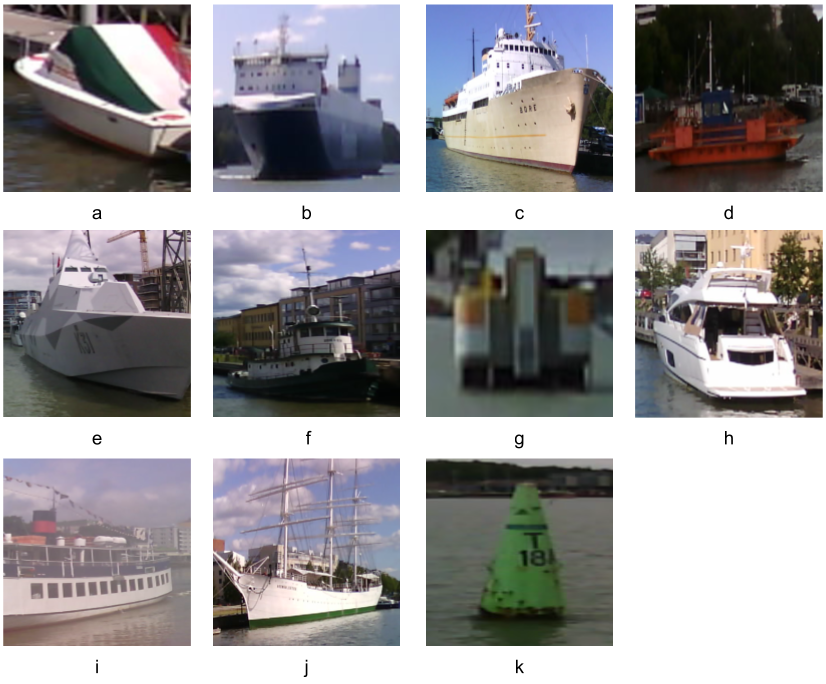}
    \caption{Example images of annotated objects in the ABOships dataset: (a) boat, (b) cargoship, (c) cruiseship, (d) ferry, (e) militaryship, (f) miscboat, (g) miscellaneous (floater), (h) motorboat, (i) passengership, (j) sailboat and (k) seamark. }
    \label{fig:label_def}
\end{figure}

\subsection{Relabelling Algorithm}
The labelling was performed by multiple people with different backgrounds, hence some label types were interpreted differently among them. To increase the consistency of labelling, we used the continuous nature of the raw data by tracking the labels between frames using the CSRT tracker \cite{lukezic2017discriminative}. 

\begin{figure}[!htb]
  \centering
    \centering
    \includegraphics[width=1\linewidth]{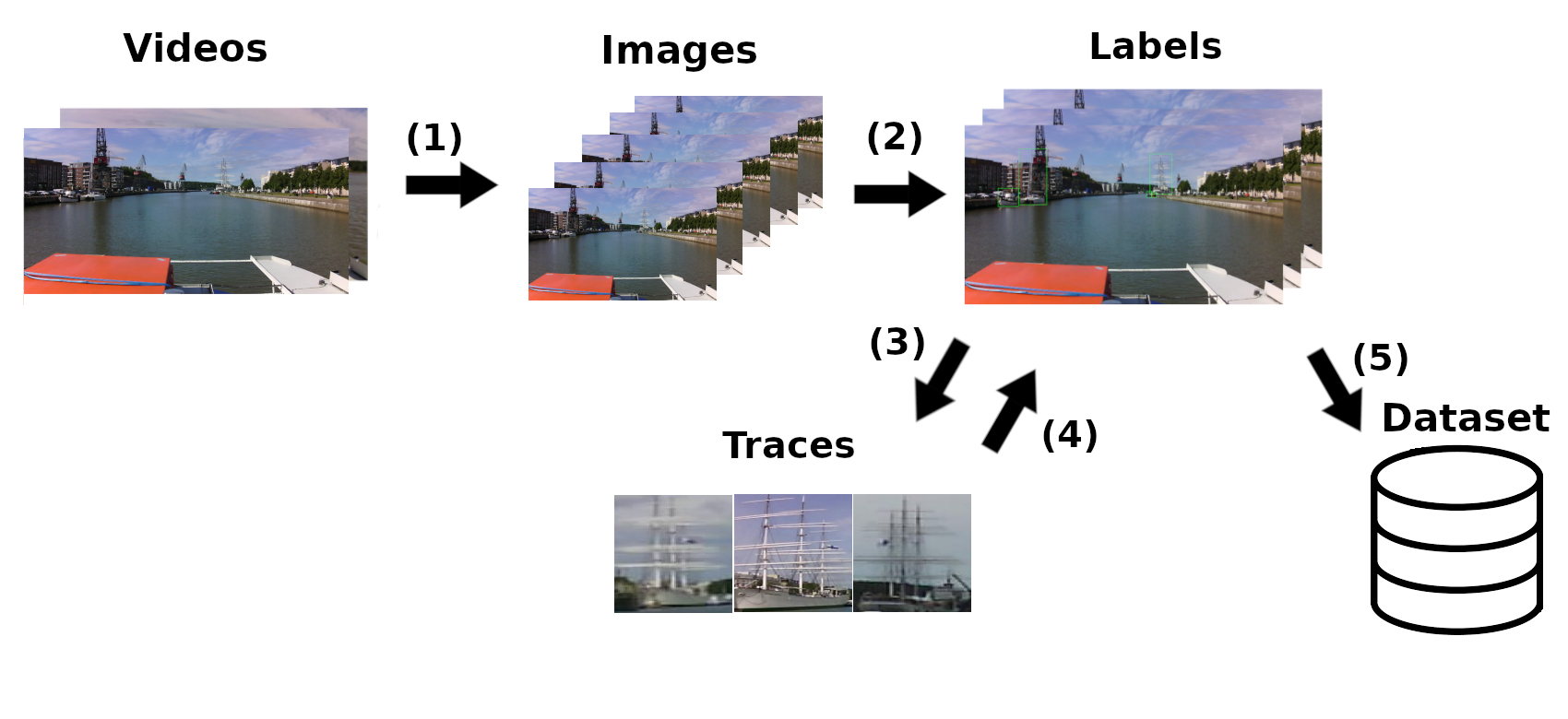}
    \caption{Videos were separated into 48 workpackages of images (1), which were labelled (2). Using the OpenCV Tracker, the objects were tracked across frames to produce traces (3) and then relabelled to fix inconsistensies and fill in the labels that might have been skipped (4). The resulting labels were then compiled into the dataset (5).}
    \label{fig:labeling_process}
\end{figure}

For every labelled frame, we created a tracker instance. The aim was to track the object until the next labelled frame. At that point, the existing traces would be mapped onto the labels on the new frame, based on the IoU metric. During this mapping, we assumed that no labeller would confuse seamarks with vessels, so we did not map ship labels onto seamarks or vice-versa. More importantly, we did not take into account previous labels, so even if annotators labelled the same object differently in different frames, these labels would still belong to the same trace as long as the tracker could identify them. For cases where the mapping could not be found, we assigned the trace a new label - "Unlabeled" to denote that even though nothing was labeled in that specific case, the tracker indicated that the tracked object should belong to the trace. A visual depiction of this process is illustrated in Figure~\ref{fig:labeling_process}.

After a certain number of frames, either the tracker would lose the object (the most common reasons for this being object occlusion, or due to the object being either too far or going out of the frame altogether) or the tracker would have no real label mapped to it enough times (which would mean it most likely drifted onto another object or the object got too far). In both of those cases, the tracker was stopped and the resulting trace  was saved to a file for further processing.

To reduce the number of errors caused by occlusion and the tracker drifting towards other objects, we performed a second tracking in the backwards direction. By comparing labels identified in the traces we acquired from tracking videos in different directions, we could detect situations where traces could not be mapped onto each other. Those cases signify that the tracker was either occluded or drifted to another object, so we had to split traces into smaller sequences until no more conflicts could be detected.

\begin{figure}[!htb]
  \centering
    \centering
    \includegraphics[width=1\linewidth]{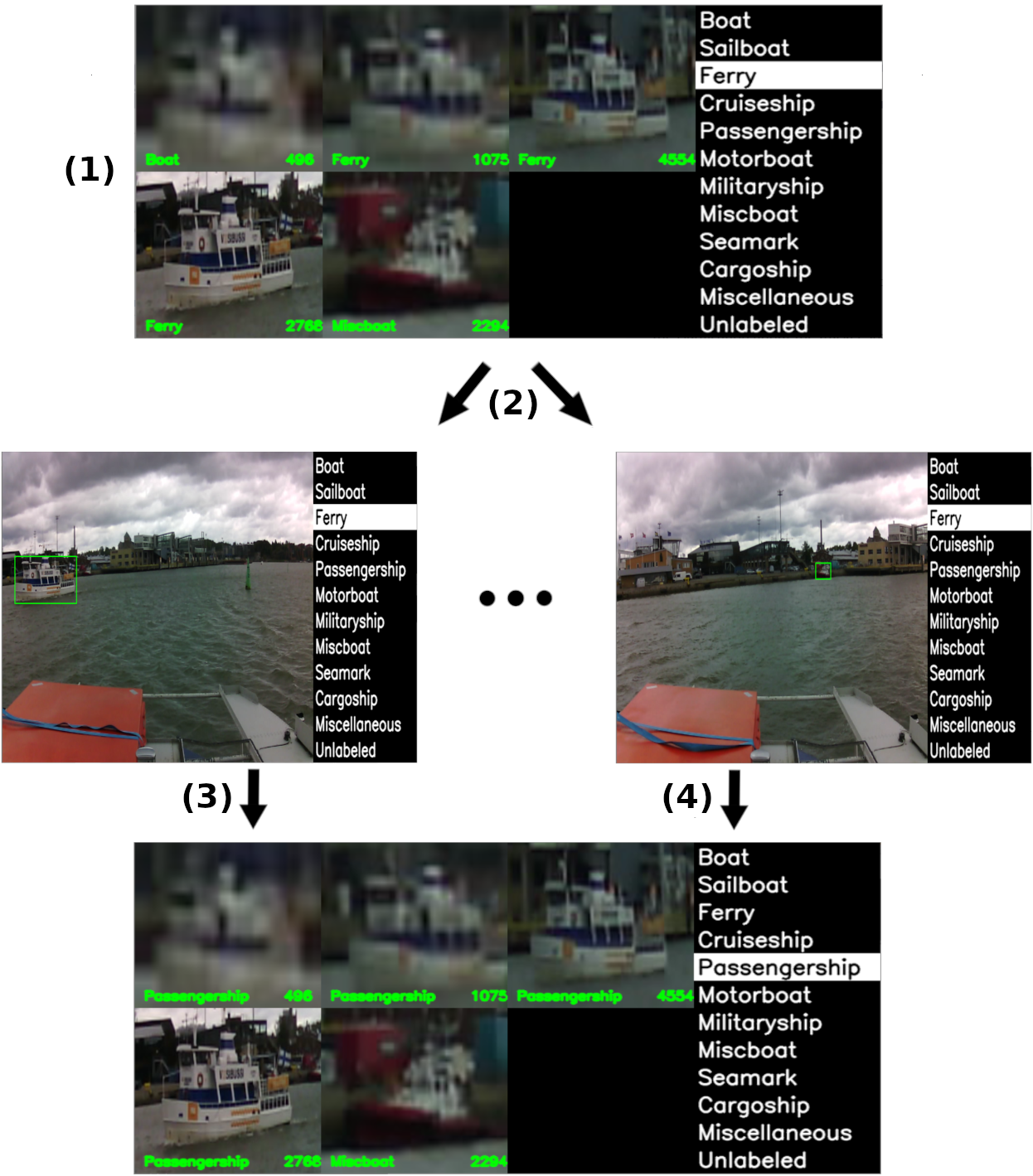}
    \caption{During the relabelling participants were given a program that would show them traces of tracked images between annotation frames (1). They were then asked to either relabel everything by selecting the correct label from the right panel, or go inside an annotation (by selecting a label that stood out (2) ) and change the label of each image individually and possibly fix the bounding box to fit the object more tightly (3). Sometimes tracking would drift to another object, in which case that particular entry of the trace might have a different label from the rest (4). When the participant completed all the labels (5), the changes would be saved into a new file and the participant was provided with the next trace.}
    \label{fig:traces}
\end{figure}

The resulting traces (after the backwards tracking) were provided as batches for relabelling. Traces containing a single entry were batched together with other singular traces from the same category. This setup was done with the purpose of preventing and removing accidental labels (mislabeling), while at the same time providing more information about the objects being labelled. This allowed us to accurately label even the objects at a longer distance with the help of tracking history. Traces obtained in this way were then provided for relabelling as a collection of labels belonging to the same trace and participants were asked to refine the labels so that labels would correspond. Singular entries that did not have a trace were batched together with other objects of the same category. The process described above is illustrated in Figure~\ref{fig:traces}.

\subsection{Dataset Statistics}
Table~\ref{tab:dataset_stat} shows the number of images of each category in our dataset and the number of annotations. The column Images represents the number of images that contain that particular object class and then the percentage of images that comprise that class follows. Then, the column Objects represents the number of annotations for that particular class in the dataset, along with the percentage of objects annotated for that specific class out of all the annotated objects in the dataset.   

\begin{figure}
\centering
    \subfloat[]{\label{small_obj}\includegraphics[width=100mm,scale=1]{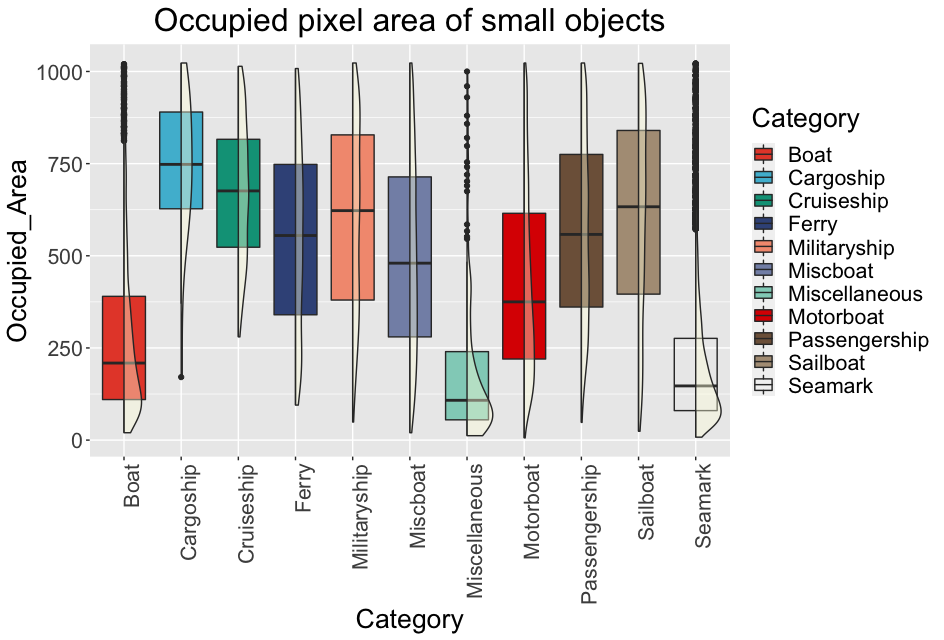}} \\
   \subfloat[]{\label{med_obj}\includegraphics[width=100mm,scale=1]{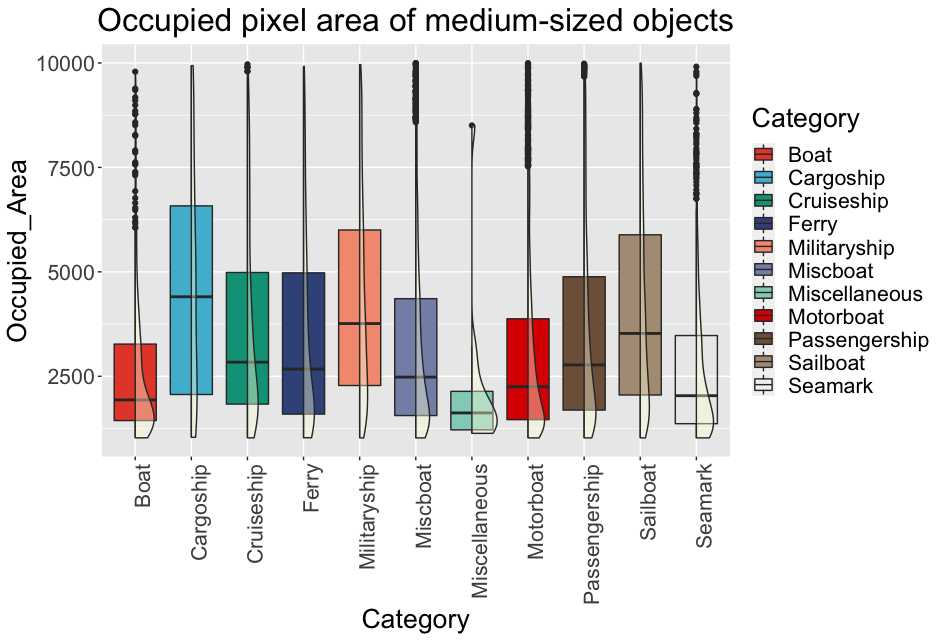}}\\
    \subfloat[]{\label{big_obj}\includegraphics[width=100mm,scale=1]{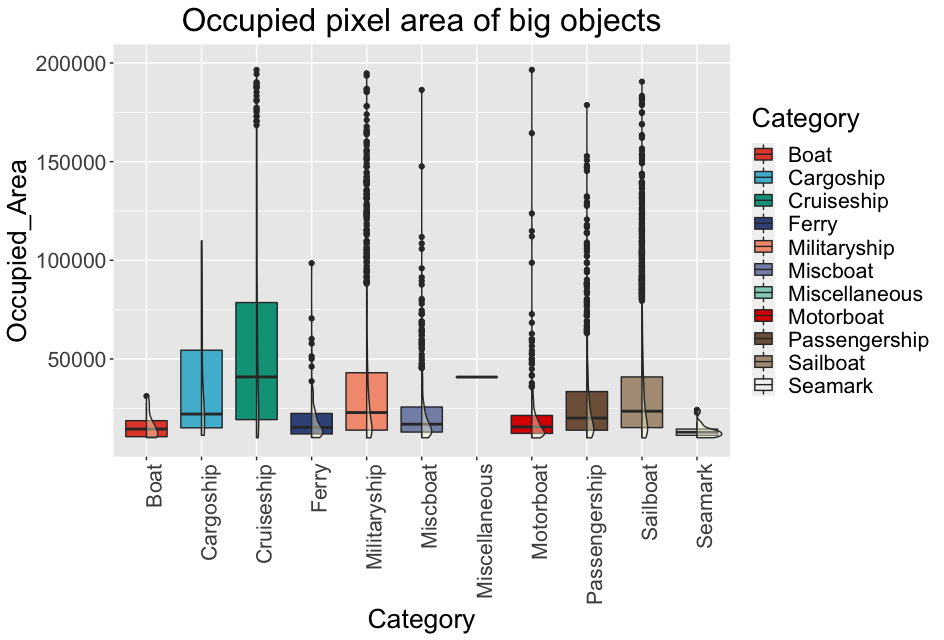}}
    \caption{Occupied pixel area of annotated objects, divided in three categories: small: Fig.~\ref{small_obj}, medium: Fig.~\ref{med_obj} and big: Fig.~\ref{big_obj}.}
    \label{stat_boxplots}
\end{figure}

Moreover, Figure~\ref{stat_boxplots} illustrates the distribution of annotated objects in our dataset based on occupied pixel area. 

\begin{table}[htb]
    \centering
    \begin{tabular}{ |p{2cm}|p{2cm}|p{2cm}|p{2cm}|p{2cm}| }

\hline
\multicolumn{5}{|c|}{Number of images and annotations for every object category} \\
\hline
Class & Images & Percentage & Objects & Percentage\\
\hline
Seamark       & 3744 & 37.89\%  & 7670 & 18.27\%  \\
Boat          & 2034 & 20.58\%  & 2913 &  6.94\%  \\
Sailboat      & 3842 & 38.88\%  & 8147 & 19.41\%  \\
Motorboat     & 4062 & 41.11\%  & 7092 & 16.89\%  \\
Passengership & 2639 & 26.71\%  & 4464 & 10.63\%  \\
Cargoship     &  157 & 1.58\%   &  161 &  0.38\%  \\
Ferry         &  945 & 9.56\%   & 1046 & 2.49\%  \\
Miscboat      & 2797 & 28.30\%  & 4642 & 11.06\%  \\
Miscellaneous &  129 & 1.30\%   & 200  &  0.47\%  \\
Militaryship  & 2559 & 25.90\%  & 4128 &  9.83\% \\
Cruiseship    & 1347 & 13.63\%  & 1504 &  3.58\% \\
\hline
\end{tabular}\bigskip
    \caption{The table shows the number of images and annotations in the ABOships dataset for every object category, along with their overall percentages.}
    \label{tab:dataset_stat}
\end{table}

\section{Experimental Results}\label{sec:basline}
\subsection{Evaluation Criteria}
To evaluate the performance of different object detection algorithms on specific datasets, one can emply various quantitative indicators. One of the most popular measures in object detection is the \textit{IoU} (\textit{Intersection of Union}), which  defines the extent of overlap of two bounding boxes as the intersection between the area of the predicted bounding-box $B_p$ and the area ground truth bounding-box $B_{gt}$, over their union~\cite{padilla2020survey}:
\begin{equation}
IoU = \frac{|B_p \cap B_{gt}|}{|B_p \cup B_{gt}|}
\end{equation}

Given an overlap threshold $t$, one can estimate whether a predicted bounding-box belongs to the background ($IoU<threshold$) or to the given classification system ($IoU>threshold$).
With this measure, one can proceed to assess the \textit{average precision} ($AP$) by calculating the precision and recall. The precision reflects the capability of a given detector to identify relevant objects and it is calculated as the proportion of detected bounding-boxes, correctly identified, over the total number of detected boxes. The recall reflects the capability of a detector to identify relevant cases and it is calculated as the proportion of correct positive predictions to all ground-truth bounding-boxes. Based on these two metrics one can draw a precision-recall curve, which encloses an area representing the average precision. However, in a majority of cases, this curve is highly irregular (zigzag pattern) making it challenging to estimate the area under it, i.e. the $AP$. To address this, one can approach it as an interpolation problem, either as an 11-point interpolation or an all-point interpolation~\cite{padilla2020survey}. 

The 11-point interpolation averages the maximum values of precision over 11 recall levels that are uniformly distributed~\cite{padilla2020survey}, as depicted below:  
\begin{equation}
AP_{11} = \sum_{R \in \{0,0.1,...,0.9,1\}} P_i(R),
\end{equation}
with 

\begin{equation}
P_i(R) = \max_{R^{*}| R^{*} \geq R} P_i(R^{*}).
\end{equation} 

$AP_{11}$ is calculated using the maximum precision $P_i(R)$, with a recall greater than $R$. 

\subsection{Baseline Detection}
To explore the performance of CNN-based object detectors on our dataset, we focused on prevalent detectors: one-stage(SSD~\cite{SSD} and EfficientDet~\cite{tan2020efficientdet}) and two-stage detectors (Faster R-CNN~\cite{ren2015faster} and R-FCN~\cite{R-FCN}). The detectors were previously
trained on the Microsoft COCO object detection dataset, which comprises a number of 91 object categories. The training dataset contains a number of $3,146$ images of marine vessels.
We investigated the performance of different feature extractors in the aforementioned detectors. We collect maritime vessel detection results based on SSD over different feature extractors (ResNet101, MobileNet v1, MobileNet v2). Moreover, we evaluate the performance of a new state-of-the-art detector, EfficientDet,  on our dataset, which used SSD-EfficientNet as feature extractor. We also evaluated two-stage detectors: Faster R-CNN and RFCN with different feature extractors. Combining all proposed detectors with the feature extractors, a total of 8 algorithm were investigated. All information regarding the specific configuration of these detectors can be found at~\cite{tensorflowAPI}. 

We estimated the performance of these algorithms in detecting maritime vessels, so we excluded seamark labels from our experiments and focused on detecting vessels. Moreover, we chose images with an occupied pixel area larger than $16^2$ pixels. We attained Table~\ref{table:results}.

Our experiments indicated that the object size impacts the detection accuracy. To corroborate this observation we divided all vessel labels (with an occupied pixel area larger than $16^2$ pixels) in our datasets into three categories, based on Microsoft COCO challenge's variants: small ($16^2$<area<$32^2$), medium ($32^2$<area<$96^2$) and large (area>$96^2$). Out of the annotated vessels with an occupied pixel area larger than $16^2$ pixels in our dataset, $30.25\%$ of the annotated vessels are small, $49.37\%$ are medium and $20.37\%$ are large.  

Analyzing the results from our experiments, we observe that detection accuracy decreases with object size. The $AP$ for best-performing detector on the dataset (Faster R-CNN with Inception ResNet v2 as feature extractor) with a registered $AP$ of $35.18\%$ more than doubles in size from small ($AP_S = 23.16\%$) to large objects ($AP_L = 46.84\%$). The second best detector on the whole dataset (EfficientDet with EfficientNet as feature extractor) however had the best performance on in the large-objects category, with an $AP_L = 55.48\%$. In general, detecting small objects turns out to be more difficult than larger object given that there is less information associated with a smaller occupied pixel area. For medium-sized objects, the best performance is attained by SSD with ResNet101 as feature extractor ($AP_M = 31.18\%$). For small objects, the best-performing detector, Faster R-CNN with Inception ResNet v2, outperforms the other detectors with a registered $AP_S = 23.16\%$. Among the SSD configurations, best performing, in general, was the one having ResNet101 as feature extractor.

\begin{table}[htb]
\centering
\begin{tabular}{ |p{2cm}|p{3cm}|p{1.5cm}|p{1.5cm}|p{1.5cm}| p{1.5cm}|}
\hline
\multicolumn{6}{|c|}{Detection performance of different  detectors on the ABOships dataset } \\
\hline
\textbf{Method} & \textbf{Feature extractor} & $AP_S$ & $AP_M$ & $AP_L$  & $AP$\\
\hline
\multirow{3}{*}{Faster RCNN} &  Inception ResNet V2 & 23.16 & 30.86 & 46.84 & \textbf{35.18} \\
\cline{2-6}
& ResNet50 V1 & 9.76 & 20.94 & 41.65 & 26.49 \\
\cline{2-6}
& ResNet101 & 18.42 & 25.07 & 38.17 & 30.26 \\
\hline
\multirow{3}{*}{SSD} &  ResNet101 V1 FPN & 21.39 & 31.18 & 42.07 & 30.03 \\
\cline{2-6}
& MobileNet V1 FPN & 12.34 & 27.61 & 37.83  & 28.59\\
\cline{2-6}
& MobileNet V2 & 3.01 & 17.05 & 27.37 & 17.48 \\
\hline
EfficientDet & Efficient Net D1 & 10.94 & 29.68 & \textbf{55.48} & 33.83\\
\hline
RFCN & ResNet101 & 18.05 & 26.20 & 41.61 & 32.46 \\
\hline
\end{tabular}\bigskip
\caption{Average Precision (AP) (in \%) of the proposed CNN-based detectors on ABOships dataset, with different feature
extractors and object sizes, for all objects with an occupied pixel area > $16^2$ pixels.}
\end{table}\label{table:results}

\begin{figure}[!htb]
  \centering
    \centering
    \includegraphics[width=1\linewidth]{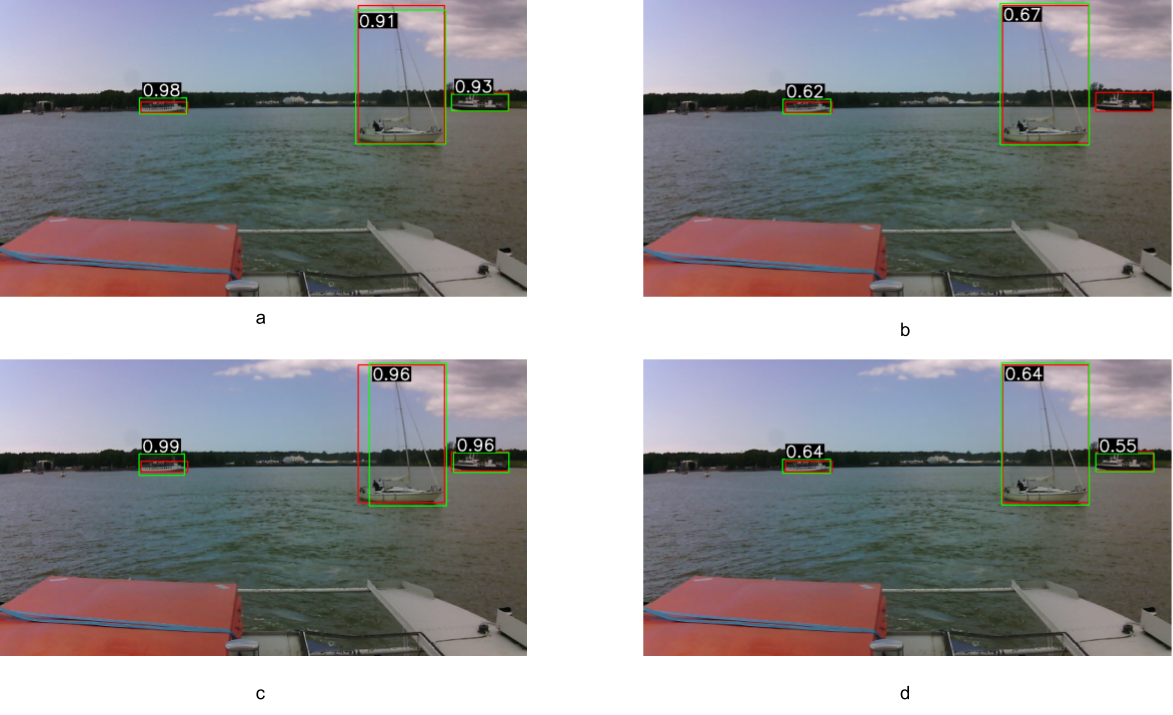}
    \caption{Qualitative detection results for the ABOships dataset on (a) Faster R-CNN and Inception-ResNet-v2 as feature extractor, (b) EfficientDet with EfficientNet as feature extractor, (c) R-FCN with ResNet101 as feature extractor, and (d) SSD with ResNet101 as feature extractor. The ground truth bounding boxes are shown as red rectangles. Predicted boxes by these methods are depicted as green bounding boxes. Each output box is associated with a class label and a score with a value in the interval [0, 1].}
    \label{fig:qualitative}
\end{figure}

\paragraph{\textbf{Qualitative results}}
Fig.~\ref{fig:qualitative} illustrates an example of detection results for the proposed methods, selecting for each the combination of feature extractor that scored the highest AP in each category. We can notice in Figure~\ref{fig:qualitative} that Faster R-CNN with a Inception-ResNet-v2 feature extractor (a) and R-FCN with a ResNet101 feature extractor (c) provide detected regions registering high scores ranging from $0.93$ to $0.99$. The other two detectors in Figure~\ref{fig:qualitative}, EfficientDet with EfficientNet as feature extractor (b) and SSD with ResNet101 as feature extractor (d), register satisfying results registering with scores ranging from $0.55$ to $0.67$. 

\newpage
\section{Conclusion}\label{sec:conclusion}
This paper provides a solution for addressing the annotation inconsistencies appeared as a consequence of manual labeling of images, using the CSRT tracker~\cite{lukezic2017discriminative}. We build traces of the images in the videos they originated from and use the CSRT tracker to traverse these videos in both directions and identify the possible inconsistencies. After this step, we employed a second round of labeling and obtained a set of $41967$ carefully annotated objects, of which $9$ types of maritime vessels (boat, miscboat, cargoship, passengership, militaryship, motorboat, ferry, cruiseship, sailboat) miscellaneous floaters and seamarks. 

We ensured the dataset consists of images taking into account the following factors: background variation, atmospheric conditions, illumination, visible proportion, occlusion and scale variation. We performed a comparison of the out-of-the-box performances of four state-of-the-art CNN-based detectors (Faster R-CNN~\cite{ren2015faster}, R-FCN~\cite{R-FCN}, SSD~\cite{SSD} and EfficientDet~\cite{tan2020efficientdet}. These detectors were previously trained on the Microsoft COCO dataset. We assess the performance of these detectors based on feature extractor and object size. Our experiments show that Faster R-CNN with Inception-Resnet v2 outperforms the other algorithms for objects with an occupied pixel area $>16^2$ pixels, except in the large object category where EfficientDet register the best performance with $AP = 55.48\%$.

For future research, we plan to investigate different types of errors in the manual labelling, for cases where the labels still have inconsistencies, such as: fine-grained recognition (which renders it more difficult for human even to detect  objects even when they are in plain view~\cite{russakovsky2015imagenet}, class unawareness (some annotators become unaware of certain classes as ground truth options) and insufficient training data (not enough training data for the annotators). 

Moreover, we plan to investigate in more detail the detection of small and very small objects, including those with an occupied pixel area $<16^2$ pixels. Also, distinguishing between different vessel types in our datasets will be an essential focus as the next steps in our experiments. In order to do this, we plan to exploit transfer learning both in the form of heterogeneous transfer learning, but also homogeneous domain adaptation. 

To further our research, we will employ maritime vessel tracking detectors on the original
videos captured in the Finnish Archipelago and examine the impact on autonomous navigation and navigational safety.

\section{Acknowledgments}
The annotation of the ABOships dataset was completed with the help of the following persons: Sabina B\"ack, Imran Shahid, Joel Sj\"oberg and Alina Torbunova.  

\bibliographystyle{unsrt}  
\bibliography{references}  

\end{document}